\documentclass[sigconf, screen]{acmart}

\usepackage{booktabs}
\usepackage{graphicx}
\usepackage{pifont}
\usepackage{mdframed}
\AtBeginDocument{%
  }

\copyrightyear{2024}
\acmYear{2024}
\setcopyright{rightsretained}
\acmConference[MMGR '24]{Proceedings of the 2nd International Workshop on Deep Multimodal Generation and Retrieval}{October 28-November 1, 2024}{Melbourne, VIC, Australia}
\acmBooktitle{Proceedings of the 2nd International Workshop on Deep Multimodal Generation and Retrieval (MMGR '24), October 28-November 1, 2024, Melbourne, VIC, Australia}
\acmDOI{10.1145/3689091.3690086}
\acmISBN{979-8-4007-1202-9/24/10}

\acmConference[MMGR '24]{Proceedings of the 2nd International Workshop on Deep Multimodal Generation and Retrieval}{October 28-November 1, 2024}{Melbourne, VIC, Australia}
\acmBooktitle{Proceedings of the 2nd International Workshop on Deep Multimodal Generation and Retrieval (MMGR '24), October 28-November 1, 2024, Melbourne, VIC, Australia}
\acmISBN{979-8-4007-1202-9/24/10}

\acmSubmissionID{mmgr1904}

\makeatletter
\gdef\@copyrightpermission{
  \begin{minipage}{0.3\columnwidth}
   \href{https://creativecommons.org/licenses/by-nc-sa/4.0/}{\includegraphics[width=0.90\textwidth]{4ACM-CC-by-nc-sa-88x31.eps}}
  \end{minipage}\hfill
  \begin{minipage}{0.7\columnwidth}
   \href{https://creativecommons.org/licenses/by-nc-sa/4.0/}{This work is licensed under a Creative Commons Attribution-NonCommercial-ShareAlike International 4.0 License.}
  \end{minipage}
  \vspace{5pt}
}
\makeatother

\begin{document}

\title[Shotluck Holmes: A Family of Efficient Small-Scale Large Language Vision Models \\ for Video Captioning and Summarization]{Shotluck Holmes: A Family of Efficient Small-Scale Large Language Vision Models for Video Captioning and Summarization}


\author{Richard Luo}
\email{richardluorl@gatech.edu}
\affiliation{%
  \institution{Georgia Institute of Technology}
  \city{Atlanta}
  \state{Georgia}
  \country{USA}
}

\author{Austin Peng}
\email{apeng39@gatech.edu}
\affiliation{%
  \institution{Georgia Institute of Technology}
  \city{Atlanta}
  \state{Georgia}
  \country{USA}
}

\author{Adithya Vasudev}
\email{avasudev8@gatech.edu}
\affiliation{%
  \institution{Georgia Institute of Technology}
  \city{Atlanta}
  \state{Georgia}
  \country{USA}
}

\author{Rishabh Jain}
\email{rjain343@gatech.edu}
\affiliation{%
  \institution{Georgia Institute of Technology}
  \city{Atlanta}
  \state{Georgia}
  \country{USA}
}

\renewcommand{\shortauthors}{Richard Luo, Austin Peng, Adithya Vasudev, and Rishabh Jain}

\begin{abstract}
    Video is an increasingly prominent and information-dense medium, yet it poses substantial challenges for language models. A typical video consists of a sequence of shorter segments, or shots, that collectively form a coherent narrative. Each shot is analogous to a word in a sentence where multiple data streams of information (such as visual and auditory data) must be processed simultaneously. Comprehension of the entire video requires not only understanding the visual-audio information of each shot but also requires that the model links the ideas between each shot to generate a larger, all-encompassing story. Despite significant progress in the field, current works often overlook videos’ more granular shot-by-shot semantic information. In this project, we propose a family of efficient large language vision models (LLVMs) to boost video summarization and captioning called Shotluck Holmes. By leveraging better pretraining and data collection strategies, we extend the abilities of existing small LLVMs from being able to understand a picture to being able to understand a sequence of frames. Specifically, we show that Shotluck Holmes achieves better performance than state-of-the-art results on the Shot2Story video captioning and summary task with significantly smaller and more computationally efficient models.
\end{abstract}



\settopmatter{printacmref=false}
\setcopyright{none}
\renewcommand\footnotetextcopyrightpermission[1]{}
\pagestyle{plain}

\maketitle

\section{Introduction}
Over the last five years, the capability for machine learning models to intake, understand, and critically reason with language and visual data has exploded, primarily due to advances in model architecture \cite{vaswani2017attention}, compute capability, and a huge increase in available data. In particular, multi-modal large language models (LLMs), powered by the revolutionary transformer architecture \cite{vaswani2017attention} have been able to achieve record-breaking understanding and reasoning capabilities on natural language and audiovisual understanding. Due to their resounding success in these fields \cite{openai2024gpt4}, LLMs have also been at the forefront in building intelligent agents to understand video. Video is incredibly complex, since it combines dynamic movement in a visual medium with aural narration, sounds, text. These four aspects tend to constructively and destructively interfere with each other over the course of a given video. As such, it remains challenging for state-of-the-art (SOTA) language models to effectively comprehend and reason off of them. The current SOTA approach, Shot2Story20K \cite{han2023shot2story20k}, proposes a landmark new benchmark dataset that combines visual and auditory signals through a three-stage model pipeline. Then, they use this custom benchmark dataset to train a custom model architecture. This approach greatly improved performance on single-shot narration captioning, multi-shot video summarization, video Q$\&$A, and video retrieval, showing that embedding multiple sources of information is essential for language models to gain an improved understanding of these multi-modal inputs.

In this paper, we take the advances presented by the Shot2Story20K paper and integrate it with one of the leading small-scale multi-modal model families: TinyLLaVA. Specifically, we show that it is sufficient to replace the final two stages of Shot2Story20K's three-stage vision-language model pipeline with TinyLLaVA, and that this replacement not only greatly reduces the memory usage, compute footprint, and latency, but also achieves SOTA performance (despite the much smaller model size) after finetuning on the Shot2Story20K dataset.

\section{Relevant Work}

\textbf{Large Multimodal Models.} \hspace{6px} Due to their impressive capabilities, large language models (LLMs) have garnered significant research interest in recent years \cite{chowdhery2023palm, chiang2023vicuna, gunasekar2023textbooks, zhang2024tinyllama, brown2020language}. This, combined with advancements in vision encoders \cite{zhai2023sigmoid, sun2023eva}, has led to some significant works in the multimodal Large Language Model field. Recent LLMs such as LLaVA \cite{liu2024visual} and InstructBLIP \cite{dai2024instructblip} leverage fine-tuning on existing LLM backbones with visual instruction tuning data to improve zero-shot performance and model alignment with human preferences. However, these models are rather large at 7B and 8.2B parameters respectively.

\textbf{TinyLLaVA.} \hspace{6px} The TinyLLaVA framework provides analysis on exploiting various small-scale LLMs for LLVMs, utilizing the same fine-tuning approach on visual instruction tuning data. Their research \cite{Zhou2024TinyLLaVAAF} finds that SigLIP \cite{zhai2023sigmoid} yields better performance than CLIP when combined with small-scale LLMs of varying parameter ranges, including TinyLlama (1.1B) \cite{zhang2024tinyllama} and Phi-2 (2.7B) \cite{li2023textbooks}. It was found that small-scale LLVMs (3.1B) can achieve better overall performance against existing 7B models including LLavA-1.5 and Qwen-VL on various evaluation benchmarks such as TextVQA \cite{singh2019towards}, SQA-I \cite{lu2022learn}, GQA \cite{hudson2019gqa}, VQAv2 \cite{goyal2017making}, MMB \cite{liu2023mmbench}, MME \cite{yin2023survey}, LLaVA-W \cite{liu2024visual}, POPE \cite{li2023evaluating}, and M-Vet \cite{yu2023mm}. However, the capabilities of TinyLLaVA are limited to just image-text-to-text generation.

\textbf{Shot2Story20K.} \hspace{6px} Shot2Story20K \cite{han2023shot2story20k} leverages HDvila100M \cite{xue2022advancing} (which contains abundant automatic speech recognition (ASR) content) and performs strict quality checking to collect a rich dataset of size 20023. Recently, the authors have even released a 134K version of the dataset, placing this benchmark shot-level dataset far beyond the capabilities of its competitors.

As mentioned, Shot2Story20K \cite{han2023shot2story20k} combines visual and auditory signals by creating a dataset through a three-stage model. (1) TransNetV2 \cite{souvcek2020transnet} separates videos into shots. (2) MiniGPT-4 \cite{zhu2023minigpt} performs video captioning of individual shots and the results are manually checked by human annotators, who also write narration captions of individual shots. (3) GPT-4 \cite{openai2024gpt4} combines all individual shot captions and ASR to perform full video summarization with human prompting.

\section{Method}

\subsection{Data-Preprocessing}

Multiple preprocessing steps were taken to adapt the video-based Shot2Story20K dataset for compatibility with TinyLLaVA. The Shotluck Holmes preprocessing procedure involved segmenting the Shot2Story20K annotations into entries comprising individual shots, each paired with its corresponding caption and ASR transcription. Subsequently, the annotations underwent processing to identify and eliminate corrupted data in order to remove any such corrupted shots from the dataset. The resultant Shotluck Holmes dataset is structured to include video inputs alongside a compilation of dialogues exchanged between a human prompt and the LLM ground truth output.

\subsection{Video to Tensor Conversion}

In order for the LLM to process the video data, the video is first converted into a tensor and then fed into a vision encoder such as SigLip \cite{zhai2023sigmoid}. To convert the video into a tensor, we experimented with two sampling methods inspired by the approach in LAVIS \cite{li2022lavis}: uniform sampling and head-tail sampling. Head-tail sampling forces the random sampling to sample an equal number of frames from the first half of the video and the second half of the video. Sampled frames are then concatenated and fed into a vision encoder.

\subsection{Shotluck Holmes Backbone}

The Shot2Story20K model architecture uses multiple different language models to assist in the video summarization or Q$\&$A tasks. By leveraging more efficient small-scale LLVMs like TinyLLaVA, we hope to achieve strong performance on single-shot video captioning and multi-shot video summarization with significantly reduced computational complexity by replacing the entire Shot2Story20K model architecture with a single LLVM model \cite{Zhou2024TinyLLaVAAF}. Because TinyLLaVA models are already pre-trained with vision encoders \cite{zhai2023sigmoid} and small-scale language models, our goal is to utilize Shot2Story20K's dataset and fine-tune the smaller LLM to extract similar performance as compared to Shot2Story's multi-step model.

\begin{figure}[!h]
    \centering
    \includegraphics[width=0.75\linewidth]{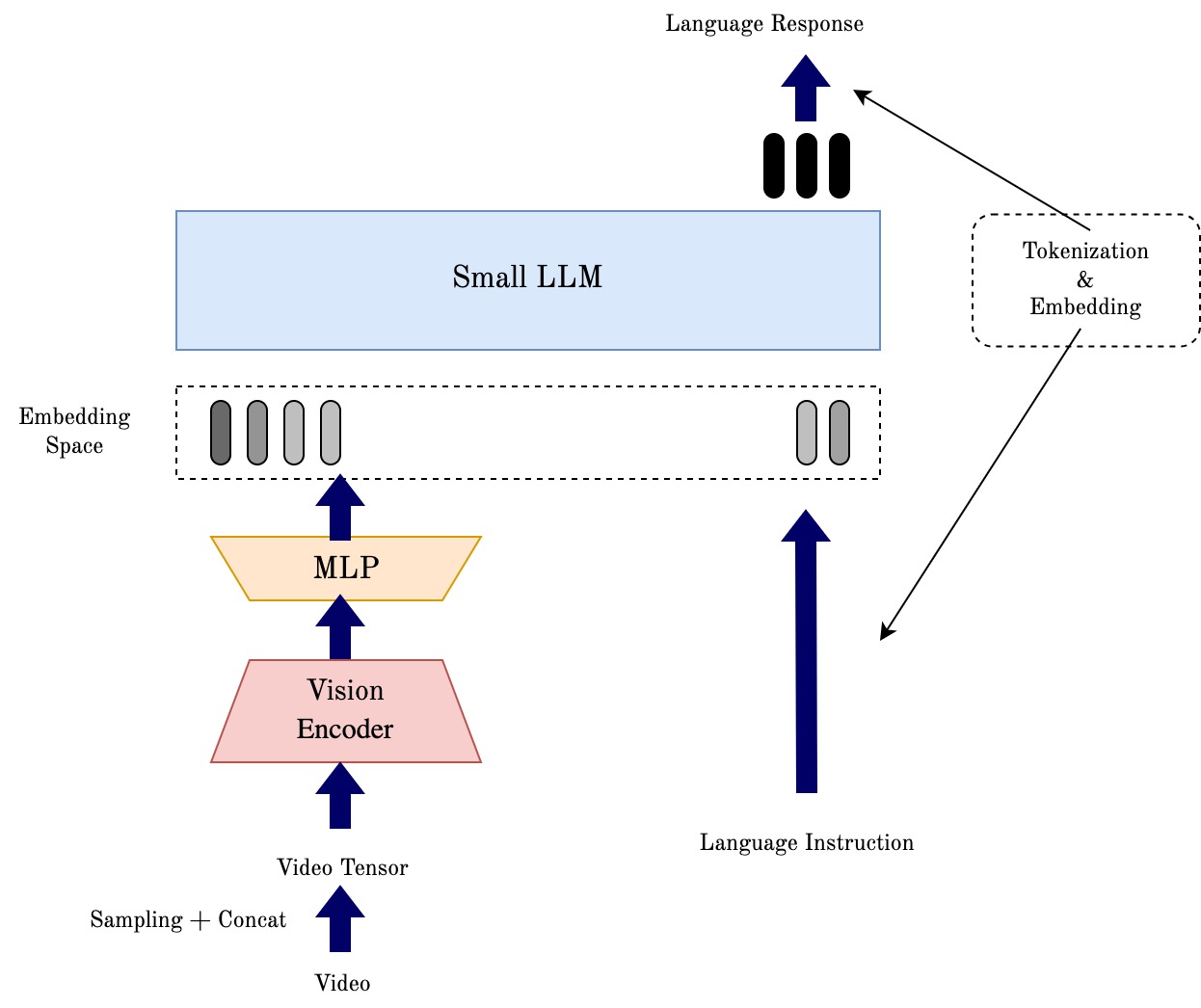}
    \caption{Shotluck Holmes model architecture \cite{Zhou2024TinyLLaVAAF}}
    \label{fig:model-arch}
\end{figure}

Shotluck Holmes presents a family of small LLMs finetuned on the Shot2Story20K dataset. We present our first two models: a 1.5B parameter LLM and a 3.1B LLM both based on TinyLLaVA, as shown in Table \ref{tab:model_size}. 

\begin{table}[htp]
  \centering
  \caption{Size of baseline models}
  \begin{tabular}{lccc}
    \toprule
    Model & Size & LLM & Vision Encoder\\
    \midrule
    Shot2Story & 7B & Vicuna & BLIP \\
    Shotluck-Holmes (1.5B) & 1.5B & TinyLlama & SigLIP \\
    Shotluck-Holmes (3.1B) & 3.1B & Phi-2 & SigLIP\\
    \bottomrule
  \end{tabular}
  \label{tab:model_size}
\end{table}

Our model architecture is shown in Figure \ref{fig:model-arch}. We follow TinyLLaVA's original pipeline of feeding in visual data (which in our case requires additional processing of the video) into the vision encoder, which is then mapped by a MLP into the LLM embedding space. During our fine-tuning, we freeze the first 12 layers of the vision encoder and update the rest of the model.

\subsection{Single-Shot Video Captioning}
The goal of single-shot video captioning is to generate descriptions for individual video shots (i.e. sections of a full video). The model first samples $N = 120$ frames from a video shot using one of two sampling methods as described earlier. We chose $N=120$ based on the largest number of frames that we could feasiblty fit into our training hardware. These frames are then concatenated and fed into the vision encoder to produce visual tokens. The tokens are fed into a MLP and then concatenated to a predetermined text prompt based on the type of LLM being used (see Appendix A) with ASR text as additional context clues. Finally, this tensor is fed through the small-scale LLM of the given size to generate the caption for the video shot.

\subsection{Multi-shot Video Summarization}

Multi-shot video summarization involves providing a rational summary describing a progression of events across different shots taken from the same video. We follow the same approach as the single-shot video captioning: we sample the same number of frames, except this time we sample from the entire video. ASR information is also retrieved from the entire video.

\section{Experiments}

For both single-shot video captioning and video summarization, we follow the same instruction template as described in Appendix A. Note that fine-tuning in all scenarios is supervised and thus includes the ground truth, which is removed during evaluation. Our optimizer and finetuning hyperparameters are listed in Appendix B.

\subsection{Single-Shot Video Captioning}
With the addition of context clues and ASR, the results are quite successful. For example, given the challenging shot below in Figure \ref{fig:frame}, the model generates the sentence: "In the video, a man in a black suit and tie is standing in front of a large screen displaying a boxing match. He is speaking into a microphone in front him."

\begin{figure}[!htp]
  \centering
  \includegraphics[width=\linewidth]{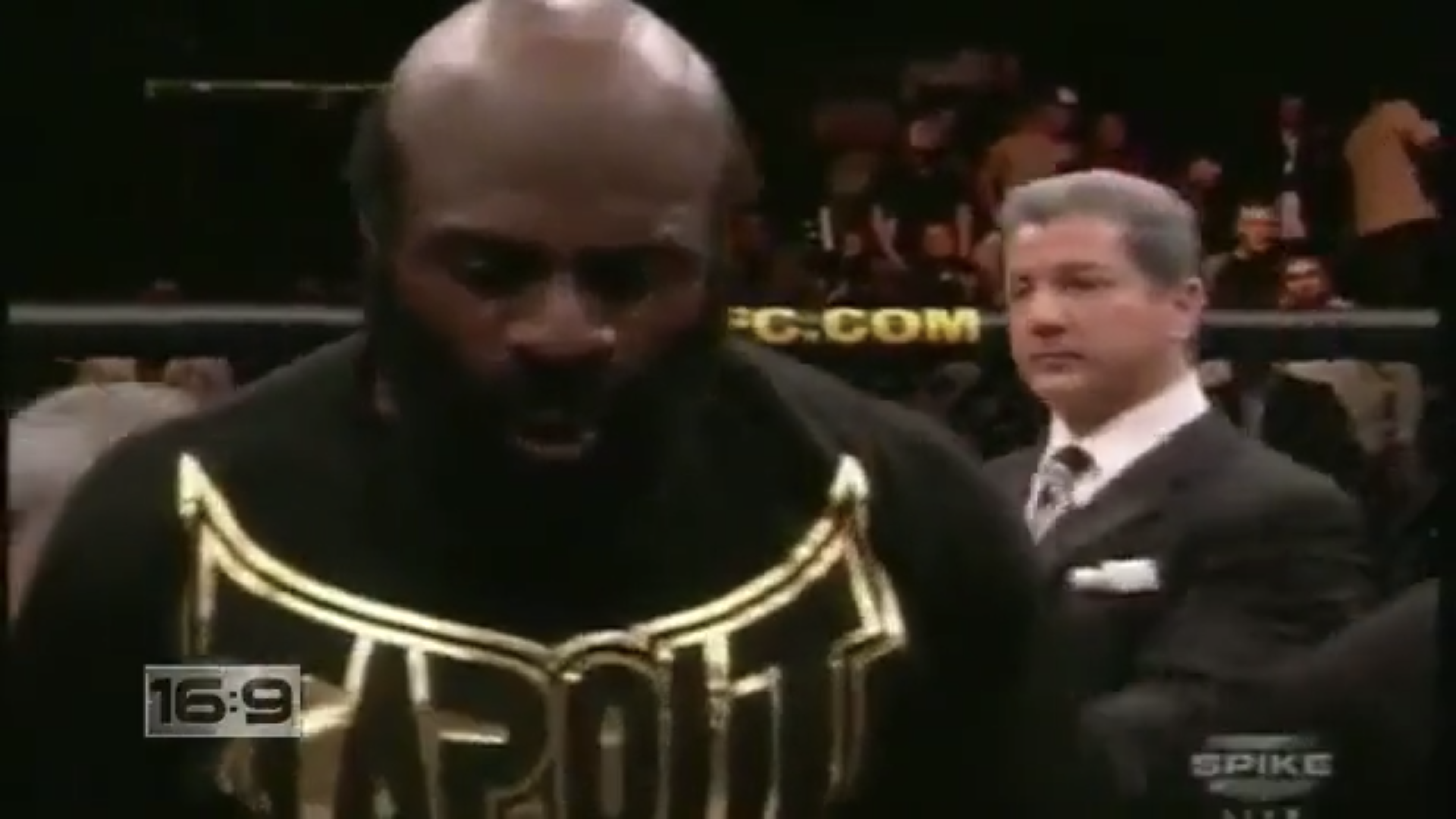}
  \caption{The sampled frame for single-shot captioning}
  \label{fig:frame}
\end{figure}

\subsection{Multi-Shot Video Summarization}

Our dataset included both single-shot and multi-shot video examples and the model was simultaneously finetuned to do both. Despite this lack of focus in finetuning, the model performs feasibly well as it is able to generate coherent summaries of multi-shot videos.

For example, for a video which contains different shots of a tablet computer and narration describing its specifications and features, the model generated the following summary: "The video is about a tablet computer that has a lot of features. It has a 64 gigabyte hard drive with two gigabytes of memory, a fast Intel Atom z3740 5F quad core data processor, and a detachable magnetic hinge that allows it to be easily connected to a keyboard. The tablet also has a latch list, which is a feature that allows you to quickly and easily engage and disengage the tablet from its keyboard".

\subsection{Evaluation}

\begin{table}[!htp]
  \centering
  \caption{Performance of best models on single-shot video captioning}
  \resizebox{\columnwidth}{!}{%
  \begin{tabular}{lcccccc}
    \toprule
    Model & BLEU & METEOR & ROUGE & CIDER \\
    \midrule
    Shot2Story (7B+) & \textbf{10.7} & 16.2 & 29.6 & 37.4 \\
    Shotluck-Holmes (3.1B) & 8.7 & \textbf{25.7}	& 36.2 & 63.2\\
    Shotluck-Holmes (1.5B) & 9.3 & 25.3 & \textbf{36.3} & \textbf{68.9}\\
    \bottomrule
  \end{tabular}
  }
\label{tab:model_baseline_cap}
\end{table}

\begin{figure}[!h]
    \centering
    \includegraphics[width=0.6\linewidth]{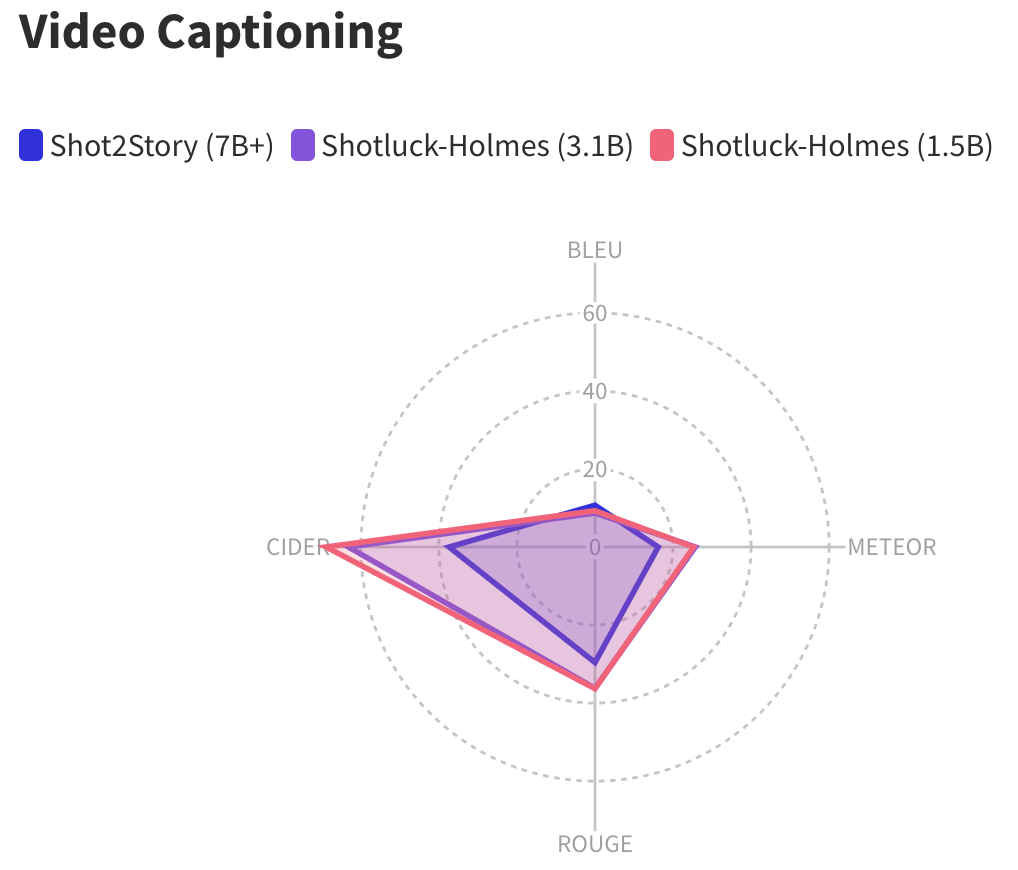}
    \caption{Single-shot video captioning results}
    \label{fig:cap-result}
\end{figure}

We evaluate our model on the task of video-captioning using BLEU@4 \citep{Papineni2002BLEU}, METEOR \citep{banerjee-lavie-2005-meteor}, and ROUGE \citep{lin-2004-rouge} on both the single-shot video captioning task and the multi-shot video summarization task. Our decoding method is top\_p sampling (see Table \ref{tab:eval_parameters} in Appendix B for numbers). We make sure to normalize our scores using the same approach as Shot2Story20K for comparability and consistency with the SOTA model's metrics. We see that on the single-shot video captioning taks, the 1.5B model of Shotluck-Holmes's 1.5B parameter model is competitive with the Shot2Story model, despite ours being around 78\% smaller. Furthermore, other than the BLEU-4 metric, which evaluates by comparing the output text sample to the baseline text sample in a manner more conducive to evaluating translations, Shotluck-Holmes 1.5B exceeds the performance of Shot2Story by between 50 and 100\%.

These performance gains are corroborated by the 3.1B model, which matches or improves on the gains seen by the 1.5B model. 

\begin{table}[!htp]
  \centering
  \caption{Performance of best models on multi-shot video summarization}
  \resizebox{\columnwidth}{!}{%
  \begin{tabular}{lcccccc}
    \toprule
    Model & BLEU & METEOR & ROUGE & CIDER \\
    \midrule
    Shot2Story (7B+) & \textbf{11.7} & 19.7 & 26.8 & 8.6 \\
    Shotluck-Holmes (3.1B) & 7.67 & \textbf{23.2} & \textbf{43} & \textbf{152.3} \\
    Shotluck-Holmes (1.5B) & 6.48 & 21.3 & 40.2 & 144.3 \\
    \bottomrule
  \end{tabular}
  }
\label{tab:model_baseline_sum}
\end{table}

\begin{figure}[!h]
    \centering
    \includegraphics[width=0.6\linewidth]{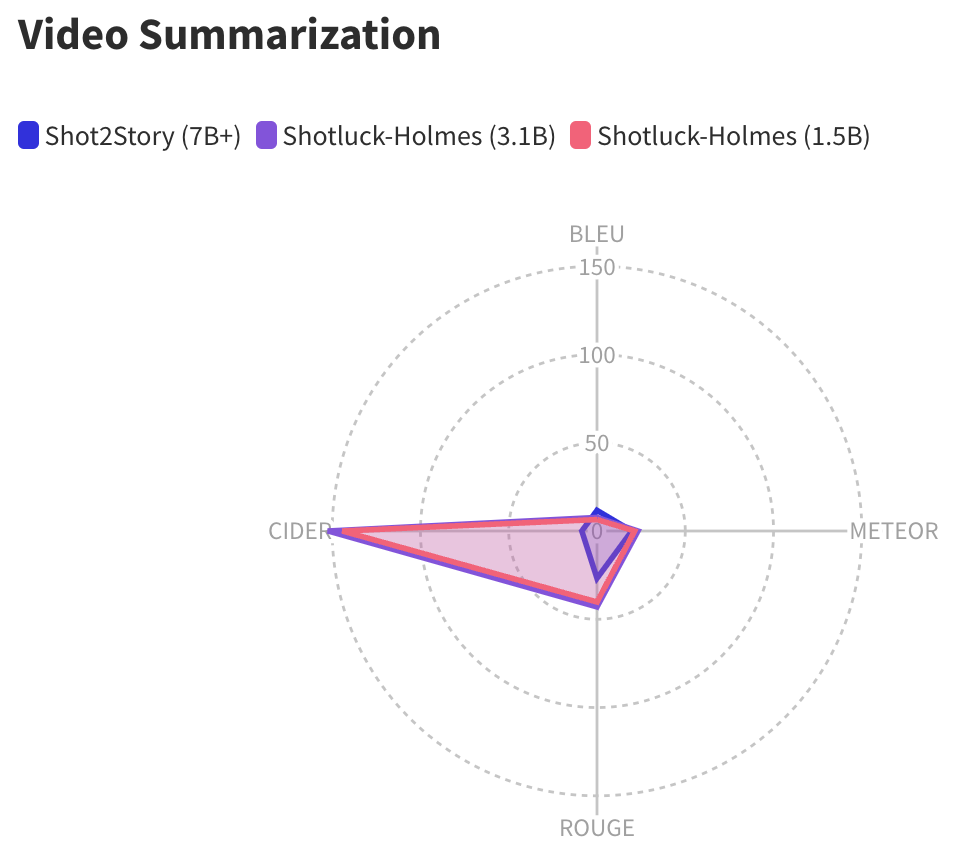}
    \caption{Multi-shot video summarization results}
    \label{fig:sum-result}
\end{figure}

We also evaluate the performance of Shot2Story vs. the two Shotluck-Holmes models on the multi-shot video summarization task using the same four metrics as mentioned above. The two models one again achieve comparable or superior performance to the larger Shot2Story model, though their improvements are slightly muted as compared to the single-shot video captioning task, achieving gains between 30 and 80\% when compared to the baseline.

\subsection{Qualitative Evaluation}

Besides systematically evaluating on public benchmarks, we further qualitatively examined how well Shotluck Holmes performed on the video captioning tasks. Shotluck Holmes 3.1B was able to summarize information chronologically from the video. It was quite accurate at transcribing the narration and did not miss any key pieces of information. In Figure \ref{fig:frame}, the model was able to infer the context of the single-shot video (a boxing match) without explicitly being told that info in a narration.

\section{Conclusion}

In this paper, we propose Shotluck-Holmes, a family of efficient models that achieve state-of-the-art performance on shot-level and full-length video understanding. This result is achieved by combining Shot2Story's \cite{han2023shot2story20k} multi-model pipeline, which integrates shot-level video annotations with audio-visual elements, with small-scale LLMs like TinyLLaVA. Furthermore, we demonstrate that these smaller models are able to achieve generalization capabilities on video captioning and summarization tasks that are competitive with larger models, suggesting that such a task is executable even on limited hardware or on edge devices. 

Despite this, there is still room for improvement with regards to the training pipeline and computational resources. One major limitation is that we trained the model on a hybrid dataset of single-shot and multi-shot videos. Although this allows for the model to generalize well across both tasks with limited samples, it prevents us from specialized fine-tuning on each particular task, which could result in even greater performance gains. 

In addition, due to compute accessibility restrictions while training and evaluating, we were forced to shard the dataset and distribute the training across multiple nodes, and it's possible that the learning rate scheduler was not set up correctly for this sharded process. 

Finally, to better represent the sequential nature of full-length video as a sequence of singular shots, a conversation feed over the shots would be required. This would allow for a proper mapping of attention over the sequence of shots. As a result, it's possible that our approach may overemphasize a single or a series of shots in a multi-shot video.

\bibliographystyle{ACM-Reference-Format}
\bibliography{sample-base}


\begin{thebibliography}{29}


\ifx \showCODEN    \undefined \def \showCODEN     #1{\unskip}     \fi
\ifx \showDOI      \undefined \def \showDOI       #1{#1}\fi
\ifx \showISBNx    \undefined \def \showISBNx     #1{\unskip}     \fi
\ifx \showISBNxiii \undefined \def \showISBNxiii  #1{\unskip}     \fi
\ifx \showISSN     \undefined \def \showISSN      #1{\unskip}     \fi
\ifx \showLCCN     \undefined \def \showLCCN      #1{\unskip}     \fi
\ifx \shownote     \undefined \def \shownote      #1{#1}          \fi
\ifx \showarticletitle \undefined \def \showarticletitle #1{#1}   \fi
\ifx \showURL      \undefined \def \showURL       {\relax}        \fi
\providecommand\bibfield[2]{#2}
\providecommand\bibinfo[2]{#2}
\providecommand\natexlab[1]{#1}
\providecommand\showeprint[2][]{arXiv:#2}

\bibitem[Banerjee and Lavie(2005)]%
        {banerjee-lavie-2005-meteor}
\bibfield{author}{\bibinfo{person}{Satanjeev Banerjee} {and} \bibinfo{person}{Alon Lavie}.} \bibinfo{year}{2005}\natexlab{}.
\newblock \showarticletitle{{METEOR}: An Automatic Metric for {MT} Evaluation with Improved Correlation with Human Judgments}. In \bibinfo{booktitle}{\emph{Proceedings of the {ACL} Workshop on Intrinsic and Extrinsic Evaluation Measures for Machine Translation and/or Summarization}}, \bibfield{editor}{\bibinfo{person}{Jade Goldstein}, \bibinfo{person}{Alon Lavie}, \bibinfo{person}{Chin-Yew Lin}, {and} \bibinfo{person}{Clare Voss}} (Eds.). \bibinfo{publisher}{Association for Computational Linguistics}, \bibinfo{address}{Ann Arbor, Michigan}, \bibinfo{pages}{65--72}.
\newblock
\urldef\tempurl%
\url{https://aclanthology.org/W05-0909}
\showURL{%
\tempurl}


\bibitem[Brown et~al\mbox{.}(2020)]%
        {brown2020language}
\bibfield{author}{\bibinfo{person}{Tom Brown}, \bibinfo{person}{Benjamin Mann}, \bibinfo{person}{Nick Ryder}, \bibinfo{person}{Melanie Subbiah}, \bibinfo{person}{Jared~D Kaplan}, \bibinfo{person}{Prafulla Dhariwal}, \bibinfo{person}{Arvind Neelakantan}, \bibinfo{person}{Pranav Shyam}, \bibinfo{person}{Girish Sastry}, \bibinfo{person}{Amanda Askell}, {et~al\mbox{.}}} \bibinfo{year}{2020}\natexlab{}.
\newblock \showarticletitle{Language models are few-shot learners}.
\newblock \bibinfo{journal}{\emph{Advances in neural information processing systems}}  \bibinfo{volume}{33} (\bibinfo{year}{2020}), \bibinfo{pages}{1877--1901}.
\newblock


\bibitem[Chiang et~al\mbox{.}({[n.\,d.]})]%
        {chiang2023vicuna}
\bibfield{author}{\bibinfo{person}{Wei-Lin Chiang}, \bibinfo{person}{Zhuohan Li}, \bibinfo{person}{Zi Lin}, \bibinfo{person}{Ying Sheng}, \bibinfo{person}{Zhanghao Wu}, \bibinfo{person}{Hao Zhang}, \bibinfo{person}{Lianmin Zheng}, \bibinfo{person}{Siyuan Zhuang}, \bibinfo{person}{Yonghao Zhuang}, \bibinfo{person}{Joseph~E Gonzalez}, {et~al\mbox{.}}} \bibinfo{year}{[n.\,d.]}\natexlab{}.
\newblock \showarticletitle{Vicuna: An open-source chatbot impressing gpt-4 with 90\%* chatgpt quality}.
\newblock  (\bibinfo{year}{[n.\,d.]}).
\newblock


\bibitem[Chowdhery et~al\mbox{.}(2023)]%
        {chowdhery2023palm}
\bibfield{author}{\bibinfo{person}{Aakanksha Chowdhery}, \bibinfo{person}{Sharan Narang}, \bibinfo{person}{Jacob Devlin}, \bibinfo{person}{Maarten Bosma}, \bibinfo{person}{Gaurav Mishra}, \bibinfo{person}{Adam Roberts}, \bibinfo{person}{Paul Barham}, \bibinfo{person}{Hyung~Won Chung}, \bibinfo{person}{Charles Sutton}, \bibinfo{person}{Sebastian Gehrmann}, {et~al\mbox{.}}} \bibinfo{year}{2023}\natexlab{}.
\newblock \showarticletitle{Palm: Scaling language modeling with pathways}.
\newblock \bibinfo{journal}{\emph{Journal of Machine Learning Research}} \bibinfo{volume}{24}, \bibinfo{number}{240} (\bibinfo{year}{2023}), \bibinfo{pages}{1--113}.
\newblock


\bibitem[Dai et~al\mbox{.}(2024)]%
        {dai2024instructblip}
\bibfield{author}{\bibinfo{person}{Wenliang Dai}, \bibinfo{person}{Junnan Li}, \bibinfo{person}{Dongxu Li}, \bibinfo{person}{Anthony Meng~Huat Tiong}, \bibinfo{person}{Junqi Zhao}, \bibinfo{person}{Weisheng Wang}, \bibinfo{person}{Boyang Li}, \bibinfo{person}{Pascale~N Fung}, {and} \bibinfo{person}{Steven Hoi}.} \bibinfo{year}{2024}\natexlab{}.
\newblock \showarticletitle{Instructblip: Towards general-purpose vision-language models with instruction tuning}.
\newblock \bibinfo{journal}{\emph{Advances in Neural Information Processing Systems}}  \bibinfo{volume}{36} (\bibinfo{year}{2024}).
\newblock


\bibitem[Goyal et~al\mbox{.}(2017)]%
        {goyal2017making}
\bibfield{author}{\bibinfo{person}{Yash Goyal}, \bibinfo{person}{Tejas Khot}, \bibinfo{person}{Douglas Summers-Stay}, \bibinfo{person}{Dhruv Batra}, {and} \bibinfo{person}{Devi Parikh}.} \bibinfo{year}{2017}\natexlab{}.
\newblock \showarticletitle{Making the v in vqa matter: Elevating the role of image understanding in visual question answering}. In \bibinfo{booktitle}{\emph{Proceedings of the IEEE conference on computer vision and pattern recognition}}. \bibinfo{pages}{6904--6913}.
\newblock


\bibitem[Gunasekar et~al\mbox{.}(2023)]%
        {gunasekar2023textbooks}
\bibfield{author}{\bibinfo{person}{Suriya Gunasekar}, \bibinfo{person}{Yi Zhang}, \bibinfo{person}{Jyoti Aneja}, \bibinfo{person}{Caio C{\'e}sar~Teodoro Mendes}, \bibinfo{person}{Allie Del~Giorno}, \bibinfo{person}{Sivakanth Gopi}, \bibinfo{person}{Mojan Javaheripi}, \bibinfo{person}{Piero Kauffmann}, \bibinfo{person}{Gustavo de Rosa}, \bibinfo{person}{Olli Saarikivi}, {et~al\mbox{.}}} \bibinfo{year}{2023}\natexlab{}.
\newblock \showarticletitle{Textbooks are all you need}.
\newblock \bibinfo{journal}{\emph{arXiv preprint arXiv:2306.11644}} (\bibinfo{year}{2023}).
\newblock


\bibitem[Han et~al\mbox{.}(2023)]%
        {han2023shot2story20k}
\bibfield{author}{\bibinfo{person}{Mingfei Han}, \bibinfo{person}{Linjie Yang}, \bibinfo{person}{Xiaojun Chang}, {and} \bibinfo{person}{Heng Wang}.} \bibinfo{year}{2023}\natexlab{}.
\newblock \showarticletitle{Shot2Story20K: A New Benchmark for Comprehensive Understanding of Multi-shot Videos}.
\newblock \bibinfo{journal}{\emph{arXiv preprint arXiv:2311.17043}} (\bibinfo{year}{2023}).
\newblock


\bibitem[Hudson and Manning(2019)]%
        {hudson2019gqa}
\bibfield{author}{\bibinfo{person}{Drew~A. Hudson} {and} \bibinfo{person}{Christopher~D. Manning}.} \bibinfo{year}{2019}\natexlab{}.
\newblock \bibinfo{title}{GQA: A New Dataset for Real-World Visual Reasoning and Compositional Question Answering}.
\newblock
\newblock
\showeprint[arxiv]{1902.09506}~[cs.CL]


\bibitem[Li et~al\mbox{.}(2022)]%
        {li2022lavis}
\bibfield{author}{\bibinfo{person}{Dongxu Li}, \bibinfo{person}{Junnan Li}, \bibinfo{person}{Hung Le}, \bibinfo{person}{Guangsen Wang}, \bibinfo{person}{Silvio Savarese}, {and} \bibinfo{person}{Steven C.~H. Hoi}.} \bibinfo{year}{2022}\natexlab{}.
\newblock \bibinfo{title}{LAVIS: A Library for Language-Vision Intelligence}.
\newblock
\newblock
\showeprint[arxiv]{2209.09019}~[cs.CV]


\bibitem[Li et~al\mbox{.}(2023a)]%
        {li2023textbooks}
\bibfield{author}{\bibinfo{person}{Yuanzhi Li}, \bibinfo{person}{S{\'e}bastien Bubeck}, \bibinfo{person}{Ronen Eldan}, \bibinfo{person}{Allie Del~Giorno}, \bibinfo{person}{Suriya Gunasekar}, {and} \bibinfo{person}{Yin~Tat Lee}.} \bibinfo{year}{2023}\natexlab{a}.
\newblock \showarticletitle{Textbooks are all you need ii: phi-1.5 technical report}.
\newblock \bibinfo{journal}{\emph{arXiv preprint arXiv:2309.05463}} (\bibinfo{year}{2023}).
\newblock


\bibitem[Li et~al\mbox{.}(2023b)]%
        {li2023evaluating}
\bibfield{author}{\bibinfo{person}{Yifan Li}, \bibinfo{person}{Yifan Du}, \bibinfo{person}{Kun Zhou}, \bibinfo{person}{Jinpeng Wang}, \bibinfo{person}{Wayne~Xin Zhao}, {and} \bibinfo{person}{Ji-Rong Wen}.} \bibinfo{year}{2023}\natexlab{b}.
\newblock \showarticletitle{Evaluating object hallucination in large vision-language models}.
\newblock \bibinfo{journal}{\emph{arXiv preprint arXiv:2305.10355}} (\bibinfo{year}{2023}).
\newblock


\bibitem[Lin(2004)]%
        {lin-2004-rouge}
\bibfield{author}{\bibinfo{person}{Chin-Yew Lin}.} \bibinfo{year}{2004}\natexlab{}.
\newblock \showarticletitle{{ROUGE}: A Package for Automatic Evaluation of Summaries}. In \bibinfo{booktitle}{\emph{Text Summarization Branches Out}}. \bibinfo{publisher}{Association for Computational Linguistics}, \bibinfo{address}{Barcelona, Spain}, \bibinfo{pages}{74--81}.
\newblock
\urldef\tempurl%
\url{https://aclanthology.org/W04-1013}
\showURL{%
\tempurl}


\bibitem[Liu et~al\mbox{.}(2024)]%
        {liu2024visual}
\bibfield{author}{\bibinfo{person}{Haotian Liu}, \bibinfo{person}{Chunyuan Li}, \bibinfo{person}{Qingyang Wu}, {and} \bibinfo{person}{Yong~Jae Lee}.} \bibinfo{year}{2024}\natexlab{}.
\newblock \showarticletitle{Visual instruction tuning}.
\newblock \bibinfo{journal}{\emph{Advances in neural information processing systems}}  \bibinfo{volume}{36} (\bibinfo{year}{2024}).
\newblock


\bibitem[Liu et~al\mbox{.}(2023)]%
        {liu2023mmbench}
\bibfield{author}{\bibinfo{person}{Yuan Liu}, \bibinfo{person}{Haodong Duan}, \bibinfo{person}{Yuanhan Zhang}, \bibinfo{person}{Bo Li}, \bibinfo{person}{Songyang Zhang}, \bibinfo{person}{Wangbo Zhao}, \bibinfo{person}{Yike Yuan}, \bibinfo{person}{Jiaqi Wang}, \bibinfo{person}{Conghui He}, \bibinfo{person}{Ziwei Liu}, {et~al\mbox{.}}} \bibinfo{year}{2023}\natexlab{}.
\newblock \showarticletitle{Mmbench: Is your multi-modal model an all-around player?}
\newblock \bibinfo{journal}{\emph{arXiv preprint arXiv:2307.06281}} (\bibinfo{year}{2023}).
\newblock


\bibitem[Lu et~al\mbox{.}(2022)]%
        {lu2022learn}
\bibfield{author}{\bibinfo{person}{Pan Lu}, \bibinfo{person}{Swaroop Mishra}, \bibinfo{person}{Tanglin Xia}, \bibinfo{person}{Liang Qiu}, \bibinfo{person}{Kai-Wei Chang}, \bibinfo{person}{Song-Chun Zhu}, \bibinfo{person}{Oyvind Tafjord}, \bibinfo{person}{Peter Clark}, {and} \bibinfo{person}{Ashwin Kalyan}.} \bibinfo{year}{2022}\natexlab{}.
\newblock \showarticletitle{Learn to explain: Multimodal reasoning via thought chains for science question answering}.
\newblock \bibinfo{journal}{\emph{Advances in Neural Information Processing Systems}}  \bibinfo{volume}{35} (\bibinfo{year}{2022}), \bibinfo{pages}{2507--2521}.
\newblock


\bibitem[OpenAI(2024)]%
        {openai2024gpt4}
\bibfield{author}{\bibinfo{person}{OpenAI}.} \bibinfo{year}{2024}\natexlab{}.
\newblock \showarticletitle{Gpt-4.}
\newblock  (\bibinfo{year}{2024}).
\newblock


\bibitem[Papineni et~al\mbox{.}(2002)]%
        {Papineni2002BLEU}
\bibfield{author}{\bibinfo{person}{Kishore Papineni}, \bibinfo{person}{Salim Roukos}, \bibinfo{person}{Todd Ward}, {and} \bibinfo{person}{Wei-Jing Zhu}.} \bibinfo{year}{2002}\natexlab{}.
\newblock \showarticletitle{BLEU: a method for automatic evaluation of machine translation}. In \bibinfo{booktitle}{\emph{Proceedings of the 40th Annual Meeting on Association for Computational Linguistics}} (Philadelphia, Pennsylvania) \emph{(\bibinfo{series}{ACL '02})}. \bibinfo{publisher}{Association for Computational Linguistics}, \bibinfo{address}{USA}, \bibinfo{pages}{311–318}.
\newblock
\urldef\tempurl%
\url{https://doi.org/10.3115/1073083.1073135}
\showDOI{\tempurl}


\bibitem[Singh et~al\mbox{.}(2019)]%
        {singh2019towards}
\bibfield{author}{\bibinfo{person}{Amanpreet Singh}, \bibinfo{person}{Vivek Natarjan}, \bibinfo{person}{Meet Shah}, \bibinfo{person}{Yu Jiang}, \bibinfo{person}{Xinlei Chen}, \bibinfo{person}{Devi Parikh}, {and} \bibinfo{person}{Marcus Rohrbach}.} \bibinfo{year}{2019}\natexlab{}.
\newblock \showarticletitle{Towards VQA Models That Can Read}. In \bibinfo{booktitle}{\emph{Proceedings of the IEEE Conference on Computer Vision and Pattern Recognition}}. \bibinfo{pages}{8317--8326}.
\newblock


\bibitem[Sou{\v{c}}ek and Loko{\v{c}}(2020)]%
        {souvcek2020transnet}
\bibfield{author}{\bibinfo{person}{Tom{\'a}{\v{s}} Sou{\v{c}}ek} {and} \bibinfo{person}{Jakub Loko{\v{c}}}.} \bibinfo{year}{2020}\natexlab{}.
\newblock \showarticletitle{Transnet v2: An effective deep network architecture for fast shot transition detection}.
\newblock \bibinfo{journal}{\emph{arXiv preprint arXiv:2008.04838}} (\bibinfo{year}{2020}).
\newblock


\bibitem[Sun et~al\mbox{.}(2023)]%
        {sun2023eva}
\bibfield{author}{\bibinfo{person}{Quan Sun}, \bibinfo{person}{Yuxin Fang}, \bibinfo{person}{Ledell Wu}, \bibinfo{person}{Xinlong Wang}, {and} \bibinfo{person}{Yue Cao}.} \bibinfo{year}{2023}\natexlab{}.
\newblock \showarticletitle{Eva-clip: Improved training techniques for clip at scale}.
\newblock \bibinfo{journal}{\emph{arXiv preprint arXiv:2303.15389}} (\bibinfo{year}{2023}).
\newblock


\bibitem[Vaswani et~al\mbox{.}(2017)]%
        {vaswani2017attention}
\bibfield{author}{\bibinfo{person}{Ashish Vaswani}, \bibinfo{person}{Noam Shazeer}, \bibinfo{person}{Niki Parmar}, \bibinfo{person}{Jakob Uszkoreit}, \bibinfo{person}{Llion Jones}, \bibinfo{person}{Aidan~N Gomez}, \bibinfo{person}{{\L}ukasz Kaiser}, {and} \bibinfo{person}{Illia Polosukhin}.} \bibinfo{year}{2017}\natexlab{}.
\newblock \showarticletitle{Attention is all you need}.
\newblock \bibinfo{journal}{\emph{Advances in neural information processing systems}}  \bibinfo{volume}{30} (\bibinfo{year}{2017}).
\newblock


\bibitem[Xue et~al\mbox{.}(2022)]%
        {xue2022advancing}
\bibfield{author}{\bibinfo{person}{Hongwei Xue}, \bibinfo{person}{Tiankai Hang}, \bibinfo{person}{Yanhong Zeng}, \bibinfo{person}{Yuchong Sun}, \bibinfo{person}{Bei Liu}, \bibinfo{person}{Huan Yang}, \bibinfo{person}{Jianlong Fu}, {and} \bibinfo{person}{Baining Guo}.} \bibinfo{year}{2022}\natexlab{}.
\newblock \showarticletitle{Advancing high-resolution video-language representation with large-scale video transcriptions}. In \bibinfo{booktitle}{\emph{Proceedings of the IEEE/CVF Conference on Computer Vision and Pattern Recognition}}. \bibinfo{pages}{5036--5045}.
\newblock


\bibitem[Yin et~al\mbox{.}(2023)]%
        {yin2023survey}
\bibfield{author}{\bibinfo{person}{Shukang Yin}, \bibinfo{person}{Chaoyou Fu}, \bibinfo{person}{Sirui Zhao}, \bibinfo{person}{Ke Li}, \bibinfo{person}{Xing Sun}, \bibinfo{person}{Tong Xu}, {and} \bibinfo{person}{Enhong Chen}.} \bibinfo{year}{2023}\natexlab{}.
\newblock \showarticletitle{A survey on multimodal large language models}.
\newblock \bibinfo{journal}{\emph{arXiv preprint arXiv:2306.13549}} (\bibinfo{year}{2023}).
\newblock


\bibitem[Yu et~al\mbox{.}(2023)]%
        {yu2023mm}
\bibfield{author}{\bibinfo{person}{Weihao Yu}, \bibinfo{person}{Zhengyuan Yang}, \bibinfo{person}{Linjie Li}, \bibinfo{person}{Jianfeng Wang}, \bibinfo{person}{Kevin Lin}, \bibinfo{person}{Zicheng Liu}, \bibinfo{person}{Xinchao Wang}, {and} \bibinfo{person}{Lijuan Wang}.} \bibinfo{year}{2023}\natexlab{}.
\newblock \showarticletitle{Mm-vet: Evaluating large multimodal models for integrated capabilities}.
\newblock \bibinfo{journal}{\emph{arXiv preprint arXiv:2308.02490}} (\bibinfo{year}{2023}).
\newblock


\bibitem[Zhai et~al\mbox{.}(2023)]%
        {zhai2023sigmoid}
\bibfield{author}{\bibinfo{person}{Xiaohua Zhai}, \bibinfo{person}{Basil Mustafa}, \bibinfo{person}{Alexander Kolesnikov}, {and} \bibinfo{person}{Lucas Beyer}.} \bibinfo{year}{2023}\natexlab{}.
\newblock \showarticletitle{Sigmoid loss for language image pre-training}. In \bibinfo{booktitle}{\emph{Proceedings of the IEEE/CVF International Conference on Computer Vision (ICCV)}}. \bibinfo{pages}{11975--11986}.
\newblock


\bibitem[Zhang et~al\mbox{.}(2024)]%
        {zhang2024tinyllama}
\bibfield{author}{\bibinfo{person}{Peiyuan Zhang}, \bibinfo{person}{Guangtao Zeng}, \bibinfo{person}{Tianduo Wang}, {and} \bibinfo{person}{Wei Lu}.} \bibinfo{year}{2024}\natexlab{}.
\newblock \showarticletitle{Tinyllama: An open-source small language model}.
\newblock \bibinfo{journal}{\emph{arXiv preprint arXiv:2401.02385}} (\bibinfo{year}{2024}).
\newblock


\bibitem[Zhou et~al\mbox{.}(2024)]%
        {Zhou2024TinyLLaVAAF}
\bibfield{author}{\bibinfo{person}{Baichuan Zhou}, \bibinfo{person}{Ying Hu}, \bibinfo{person}{Xi Weng}, \bibinfo{person}{Junlong Jia}, \bibinfo{person}{Jie Luo}, \bibinfo{person}{Xien Liu}, \bibinfo{person}{Ji Wu}, {and} \bibinfo{person}{Lei Huang}.} \bibinfo{year}{2024}\natexlab{}.
\newblock \showarticletitle{TinyLLaVA: A Framework of Small-scale Large Multimodal Models}.
\newblock \bibinfo{journal}{\emph{ArXiv}}  \bibinfo{volume}{abs/2402.14289} (\bibinfo{year}{2024}).
\newblock
\urldef\tempurl%
\url{https://api.semanticscholar.org/CorpusID:267782659}
\showURL{%
\tempurl}


\bibitem[Zhu et~al\mbox{.}(2023)]%
        {zhu2023minigpt}
\bibfield{author}{\bibinfo{person}{Deyao Zhu}, \bibinfo{person}{Jun Chen}, \bibinfo{person}{Xiaoqian Shen}, \bibinfo{person}{Xiang Li}, {and} \bibinfo{person}{Mohamed Elhoseiny}.} \bibinfo{year}{2023}\natexlab{}.
\newblock \showarticletitle{Minigpt-4: Enhancing vision-language understanding with advanced large language models}.
\newblock \bibinfo{journal}{\emph{arXiv preprint arXiv:2304.10592}} (\bibinfo{year}{2023}).
\newblock


\end{thebibliography}

\pagebreak

\appendix

\section{Instruction Templates}
\label{sec:instruction-templates}

This is our instruction template for prompting. During supervised training, we include the ground truth in assistant, but for evaluation, this ground truth is removed.

\begin{mdframed}
    A chat between a curious user and an artificial intelligence assistant. The assistant gives helpful, detailed, and polite answers to the user's questions. USER: $<Video>$

    Please describe this video. Do not include details that you are not sure of. This is what the speech in the video is saying: $<ASR>$ ASSISTANT: $<Ground Truth>$
\end{mdframed}

\section{Hyperparameters}
\label{sec:hyperparameters}

\begin{table}[!htp]
\centering
\caption{Hyperparameters for Shotluck Holmes training}
\resizebox{\columnwidth}{!}{%
\begin{tabular}{lcccccc}
\toprule
Optimizer & Global Batch Size & Learning Rate & Epochs & Max Length & Weight Decay \\
\midrule
Adam8bit & 128               & 2e-5          & 1      & 3072       & 0            \\
\bottomrule
\end{tabular}
}
\label{tab:hyperparameters}
\end{table}

\begin{table}[H]
\centering
\caption{Evaluation Parameter Settings}
\begin{tabular}{lc@{}}
\toprule
Parameter & Value \\
\midrule
temperature & 0.2 \\
top\_p & 0.9 \\
no\_repeat\_ngram\_size & 3 \\
\bottomrule
\end{tabular}
\label{tab:eval_parameters}
\end{table}

\section{Compute Resources}
\label{sec:compute-resources}
All models were trained in an environment with 2 TB of RAM, 8x NVIDIA H100 GPUs, and 64 CPU cores. Fine tuning the 1.5B parameter model took approximately 6 hours and the 3.1B parameter model 8 hours with this setup.

\section{Broader Impact}
\label{sec:broader-impact}
Shotluck Holmes is based entirely on existing LLMs and datasets and introduces no new architectures or data. As such, our work inherits any existing limitations of LLMs and the Shot2Story20K dataset, including but not limited to hallucination and biased outputs. However, our work does not introduce any new implications for societal impact or require any new safeguards. Improved video summarization capabilities do not pose any greater risk of misuse than current technologies. 

\end{document}